\documentclass{article}
\usepackage{spconf,amsmath,graphicx}
\usepackage{color}
\usepackage{multirow}
\usepackage{makecell}
\usepackage{times}
\usepackage{epsfig}
\usepackage{graphicx}
\usepackage{amsmath}
\usepackage{amssymb}
\usepackage{booktabs}
\usepackage{subcaption} 
\usepackage{enumitem}
\usepackage[breaklinks=true,bookmarks=false]{hyperref}

\title{Which country is this picture from? New data and methods for DNN-based country recognition}
\name{Omran Alamayreh, Giovanna Maria Dimitri, Jun Wang, Benedetta Tondi, Mauro Barni
\thanks{This work has been partially supported by the
		PREMIER project under contract PRIN 2017 2017Z595XS-001, funded by the Italian Ministry of University and Research, and by the Defense Advanced Research Projects Agency (DARPA) and the Air Force Research Laboratory (AFRL) under agreement number FA8750-20-2-1004. \OMRAN{I have shorten the text, The U.S. Government is authorized to}
}}
\address{Department of Information Engineering and Mathematics, University of Siena}

\begin{document}
	
	\sloppy
	\definecolor{black}{rgb}{0.0, 0.0, 0.0}
	\definecolor{white}{rgb}{1.0, 1.0, 1.0}
	\definecolor{yellow}{rgb}{1.0, 1.0, 0.8}
	\definecolor{red}{rgb}{0.6, 0.0, 0.2}
	\definecolor{blue}{rgb}{0.0, 0.2, 0.5}
	\definecolor{green}{rgb}{0.6, 0.8, 0.8}
	\definecolor{dark_green}{RGB} {0, 140, 0}
	\definecolor{gold}{rgb}{0.6, 0.4, 0.1}
	\definecolor{grey}{RGB}{0,0,0}
	\definecolor{Gray}{gray}{0.8}
	\definecolor{MediumGray}{gray}{0.9}
	\definecolor{LightGray}{gray}{0.98}
	\definecolor{LightCyan}{rgb}{0.88,1,1}
	\definecolor{purple}{RGB}{128,0,128}
	\definecolor{sl_blue}{RGB}{47, 60, 105}
	\definecolor{orange}{RGB}{255,165,0}
	\definecolor{Gray}{gray}{0.85}
	
\newif\ifcomments

\commentsfalse 

\ifcomments
	\newcommand{\JUN}[1]{\textcolor{orange}{\textbf{\small [}\colorbox{yellow}{\textbf{Jun:}}{\small #1}\textbf{\small ]}}}
	\newcommand{\GIOVANNA}[1]{\textcolor{red}{\textbf{\small [}\colorbox{yellow}{\textbf{GIOVANNA:}}{\small #1}\textbf{\small ]}}}
	\newcommand{\MB}[1]{\textcolor{dark_green}{\textbf{\small [}\colorbox{yellow}{\textbf{Mauro:}}{\small #1}\textbf{\small ]}}}
	\newcommand{\OMRAN}[1]{\textcolor{blue}{\textbf{\small [}\colorbox{yellow}{\textbf{Omran:}}{\small #1}\textbf{\small ]}}}
	\newcommand{\BTcomm}[1]{\textcolor{purple}{\textbf{\small [}\colorbox{yellow}{\textbf{Benedetta:}}{\small #1}\textbf{\small ]}}}
	\newcommand{\BT}[1]{\textcolor{brown}{{#1}}}
	\newcommand{\TODO}[1]{\textcolor{red}{{TODO: #1}}}
	\newcommand{\CH}[1]{\textcolor{purple}{{#1}}}
\else
	\newcommand{\JUN}[1]{}
	\newcommand{\GIOVANNA}[1]{}
	\newcommand{\MB}[1]{}
	\newcommand{\OMRAN}[1]{}
	\newcommand{\BTcomm}[1]{}
	\newcommand{\BT}[1]{}
	\newcommand{\TODO}[1]{}
	\newcommand{\CH}[1]{}
\fi

\ninept
	\maketitle
\begin{abstract}

Recognizing the country where a picture has been taken has many potential applications, such as identification of fake news and prevention of disinformation campaigns. Previous works focused on the estimation of the geo-coordinates where a picture has been taken. Yet, recognizing in which country an image was taken could be more critical, from a semantic and forensic point of view, than estimating its spatial coordinates. In the above framework, this paper provides two contributions. First, we introduce the VIPPGeo dataset, containing 3.8 million geo-tagged images. Secondly, we used the dataset to train a model casting the country recognition problem as a classification problem. The experiments show that our model provides better results than the current state of the art. Notably, we found that asking the network to identify the country provides better results than estimating the geo-coordinates and then tracing them back to the country where the picture was taken.
\end{abstract}
\begin{keywords}
	Image Geolocalization, Country Recognition, Fake News Detection
\end{keywords}
\vspace{-0.05cm}	
\section{Introduction}
\label{sec:intro}
\vspace{-0.2cm}

Image-based geo-localization is receiving increasing attention, 
due to its importance in many applications \cite{garg2021your} \cite{zhang2021visual}. For instance, geo-localizing the position where an image has been shot, could be particularly relevant for fact checking, to prevent the diffusion of fake news and to fight misinformation campaigns. Furthermore, detecting the exact place portrayed by a photo, could be extremely useful to identify text image inconsistencies in the news, thus revealing if a certain image was actually taken from a different location with respect to the one referred to in the text \cite{ghai2021deep} \cite{muller2018geolocation}. 

Most of the works carried out so far aim at estimating the geo-coordinates of the image scene with the best possible accuracy \cite{muller2018geolocation} \cite{workman2015wide}. In this work, we propose to adopt a different perspective, i.e. to identify the country where an image has been taken. In most cases, recognizing the country is more important (and possibly easier) than providing a precise estimate of the geo-coordinates of the scene framed by the picture. One may argue that once the geo-coordinates have been estimated, tracing back to the country is a trivial task. In fact, this is true only in an ideal case, however, in non-ideal conditions, a system which is asked to minimize the spatial localization error, may result in a wrong classification of the country, when this allows to reduce the localization error. On the other hand, being a human construction, countries (usually) define geographical regions with similar human-related features, like historical landmarks, language, architecture, social habits, and such characteristics may be conveniently used to identify the country a certain image belongs to. The contribution of this paper can be summarised as:

\begin{enumerate}[leftmargin=\parindent,align=left,labelwidth=\parindent,labelsep=0pt]
\item[1] {We present a novel dataset, named VIPPGeo, to be used for the country recognition task. The dataset contains 3.8 million geo-tagged images downloaded from the three public databases of Flickr~\cite{flickr}, Mapillary~\cite{mapillary} and Unsplash~\cite{unsplash}. The images were crawled according to a list of 243 countries to obtain a comprehensive representation of the whole set of countries in the world. The photos contain urban outdoor scenes only, due to the relevance of such kind of data for the problem at hand, in particular for detecting the architectural and semantic differences between countries. Data filtering strategies were applied to retain only relevant images, discarding, for instance, natural scene images, indoor-images, grey-level images, and images mostly occupied by specific, non-relevant, subjects for instance ships or airplanes.}
\item[2] {We use the new dataset to train a classifier based on a ResNet-101 network~\cite{he2016deep} to distinguish among countries. In order to take into account the presence of very small countries and countries with few sample images in the dataset, we grouped the countries into 61 classes. The results we got on a test subset extracted from VIPPGeo, show a significant improvement of the performance with respect to a state-of-the-art baseline method ~\cite{muller2018geolocation} that recovers the country from the estimated geo-spatial coordinates.}
\end{enumerate}


\vspace{-0.5cm}
\section{Related Works}
\label{sec:SoA}
\vspace{-0.2cm}


Geo-localization is a hard task that for its practical relevance has been the subject of a vast amount of research.


In~\cite{Planet} and~\cite{seo2018cplanet}, the authors tackle the photo geolocation problem subdividing the surface of the earth into thousands of multi-scale geographic cells, and training a deep network using millions of geotagged images mined from all over the Internet. In~\cite{Planet}, the dataset consists of non-urban images, containing indoor, portraits, pets, food products and other photos not indicative of location, while in~\cite{seo2018cplanet} the authors have preprocessed the dataset removing non-natural images (e.g., clipart images, product photos, etc.). Another interesting work is~\cite{im2gps_2}, where the authors proposed to combine DL classification with
the approach proposed in Im2GPS~\cite{im2gps}, in which a query image is matched against a database of geotagged images and the location is inferred from the retrieved set. In both works~\cite{im2gps,im2gps_2}, the dataset used in the training consists of more than 6 million images collected from Flickr in 2008 without filtering irrelevant images.



The goal of most of the methods proposed so far consists in geo-coordinates estimation, the final goal being to minimize the distance between the estimated and true coordinates. In this sense, one of the best performing systems proposed so far is described in~\cite{muller2018geolocation}. In this work, the authors propose a classification system, in which the earth is subdivided into geographical cells. Photographs taken in different types of environments (urban outdoor, indoor or natural) are considered, to embed in the learning process specific features of several environmental settings. The proposed DL architecture is based on a ResNet-50 network. Results obtained on the 2 benchmark datasets (which we also used in our experiments) show the capability of this system to outperform prior works, thus qualifying~\cite{muller2018geolocation} as the reference benchmark for geo-localization.


\begin{figure}
\centering
\begin{subfigure}{0.15\textwidth}
    \includegraphics[width=\textwidth]{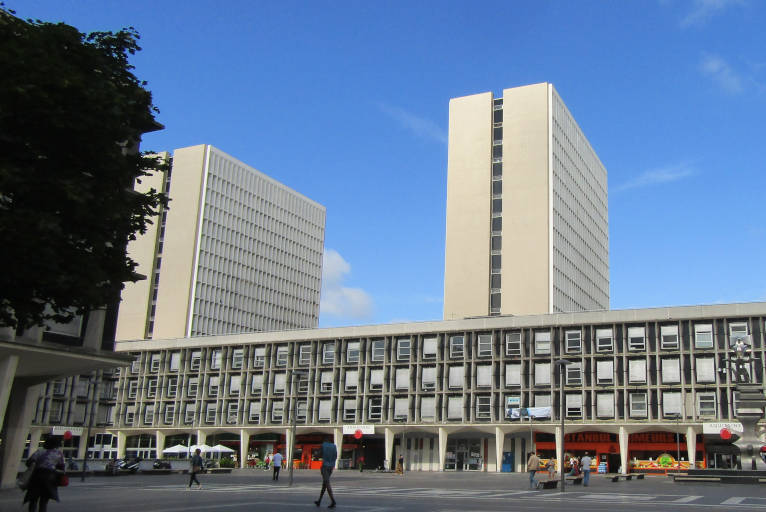}
    \caption{Flickr - France.}
    \label{fig:first}
\end{subfigure}
\hfill
\begin{subfigure}{0.15\textwidth}
    \includegraphics[width=\textwidth]{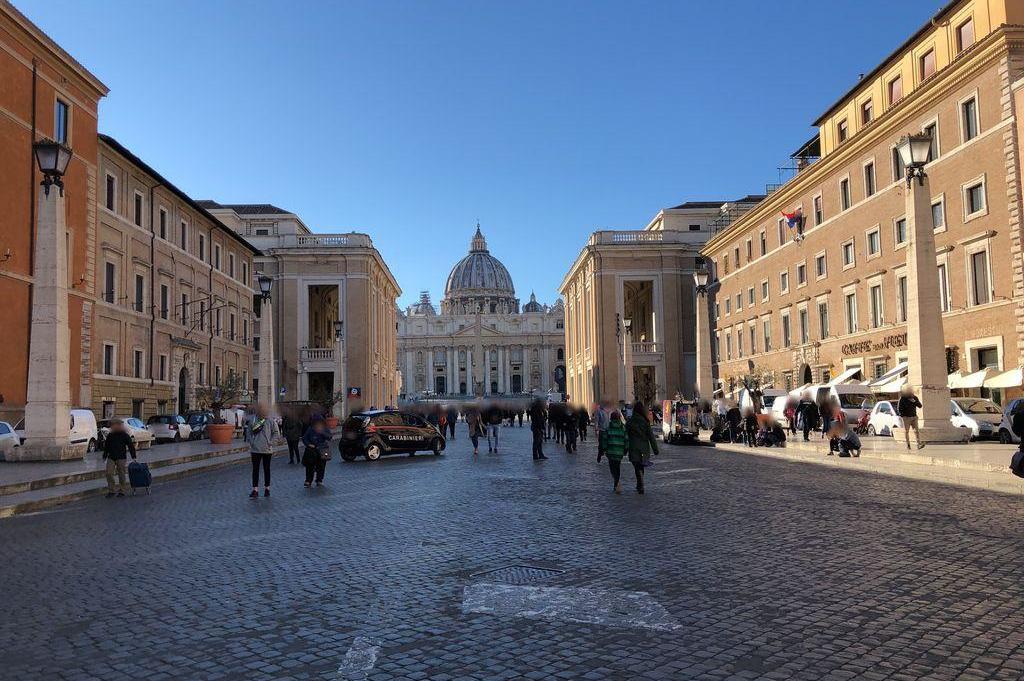}
    \caption{Mapillary - Vatican.}
    \label{fig:second}
\end{subfigure}
\hfill
\begin{subfigure}{0.15\textwidth}
    \includegraphics[width=\textwidth]{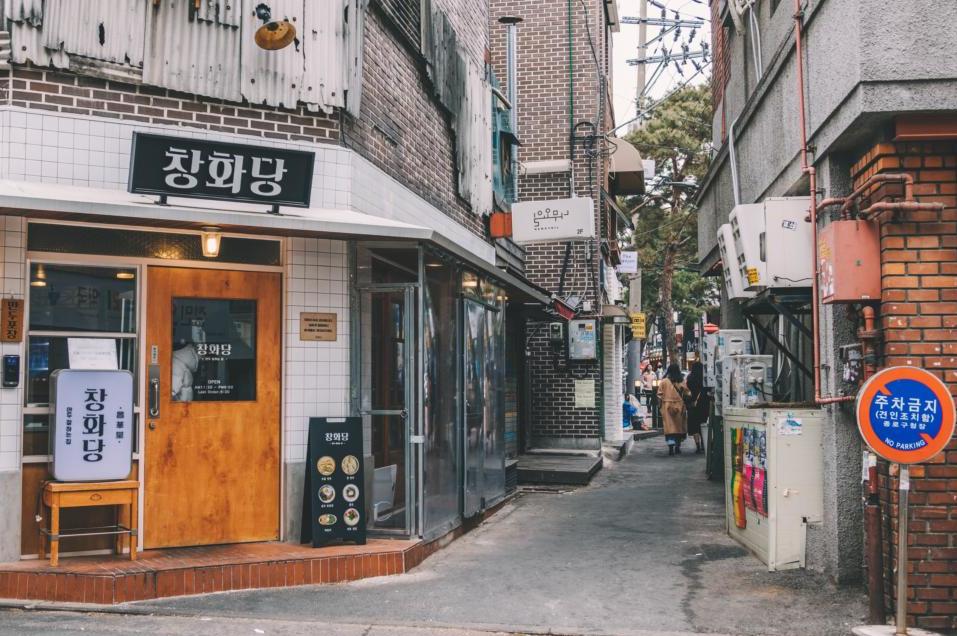}
    \caption{Unsplash - S.Korea.}
    \label{fig:third}
\end{subfigure}
    \vspace{-0.25cm}        
\caption{Three sample photos taken from different databases.}
    \vspace{-0.5cm}
\label{fig:threeDatasets_Examples}
\end{figure}


An important aspect to be taken into account is the availability of reference benchmark datasets. Developing and testing DNN-based systems for image geo-localization, in fact, requires the availability of large amounts of data capturing the differences (architectural, stylistic, social, etc \dots) among world-countries. On this regard, datasets containing geo-tagged images are rarely available. The only available dataset that we could exploit in our work is a subset of the YFCC100M dataset~\cite{thomee2016yfcc100m} containing about 5 million images; the subset is also used in the~\cite{muller2018geolocation} work. However, the subset contains many non-urban images such as landscapes and natural scenes, and indoor images, which are less relevant for the country recognition task addressed in our work. For this reason, we built and released the VIPPGeo dataset (Section~\ref{sec:dataset}).



\vspace{-0.25cm}
\section{The VIPPGeo Dataset}
\label{sec:dataset}
\vspace{-0.2cm}
We built the VIPPGeo dataset by using publicly available sources, namely: Flickr, Mapillary and Unplash. In this way, we gathered photos with different characteristics, taken with different cameras, from a wide range of different photographers and largely diverse views of all world-countries. The images were crawled using the APIs released by each of the three data-sources.

\textit{Flickr}~\cite{flickr} is a public repository where users can upload and share their personal photos. Photos uploaded on the website can have several different types of copyrights. To build our dataset, we only retrieved photos which are freely usable (no copyright associated) and made available under the Creative Commons license.


To build our dataset, we used the YFCC100M dataset \cite{thomee2016yfcc100m} containing about 5 million geo-tagged images. In addition, we have retrieved photos from Flickr which are relevant to our task. We have used search queries containing 183 Urban landmarks keywords, like ``university'', ``museum'', ``skyline'' etc \dots. The keywords were paired with all the cities/towns in the world with a population over 1000 people~\cite{geonames} for a total of 144,563 cities. For instance we used queries such as ``London church''. In total, we sent: $ 183 \text{ keys} \times 144,563 \text{ cities}=26,455,029 \text{ queries}, $ from which we collected around one million images. We then merged these images with the YFCC100M dataset~\cite{thomee2016yfcc100m} gathering about 6 million images.

 
\textit{Mapillary}~\cite{mapillary} is a web platform allowing crowd-sourcing geo-tagged images. To retrieve the urban images that we have included in our dataset, we have iterated our queries over each city/town in the world with more than 1,000 inhabitants~\cite{geonames}. Then, we requested the Mapillary API to return all images in a bounding box of $10$ kilometers around the GPS location of each city. In this way, we obtained a total of 2,528,328 images.

\textit{Unsplash}~\cite{unsplash} is a website devoted to
photo-sharing. The site has a growing database of more than 3 million photos. Unsplash has been cited as one of the world's leading photography websites, that provides high-quality images. In our dataset, we crawled the images from the publicly available `Full Dataset' database.

All images of the VIPPGeo dataset are stored in JPEG format with a quality factor equal to 75. With regard to the image size, when the smallest dimension of an image was larger than 640, we resized it so that the smallest dimension is equal to 640, otherwise, we stored the images in their original dimension. In Figure~\ref{fig:threeDatasets_Examples}, we show three examples of images from Flickr, Mapillary and Unsplash.

\begin{figure}[t!]
\centering
    \includegraphics[scale=0.15]{"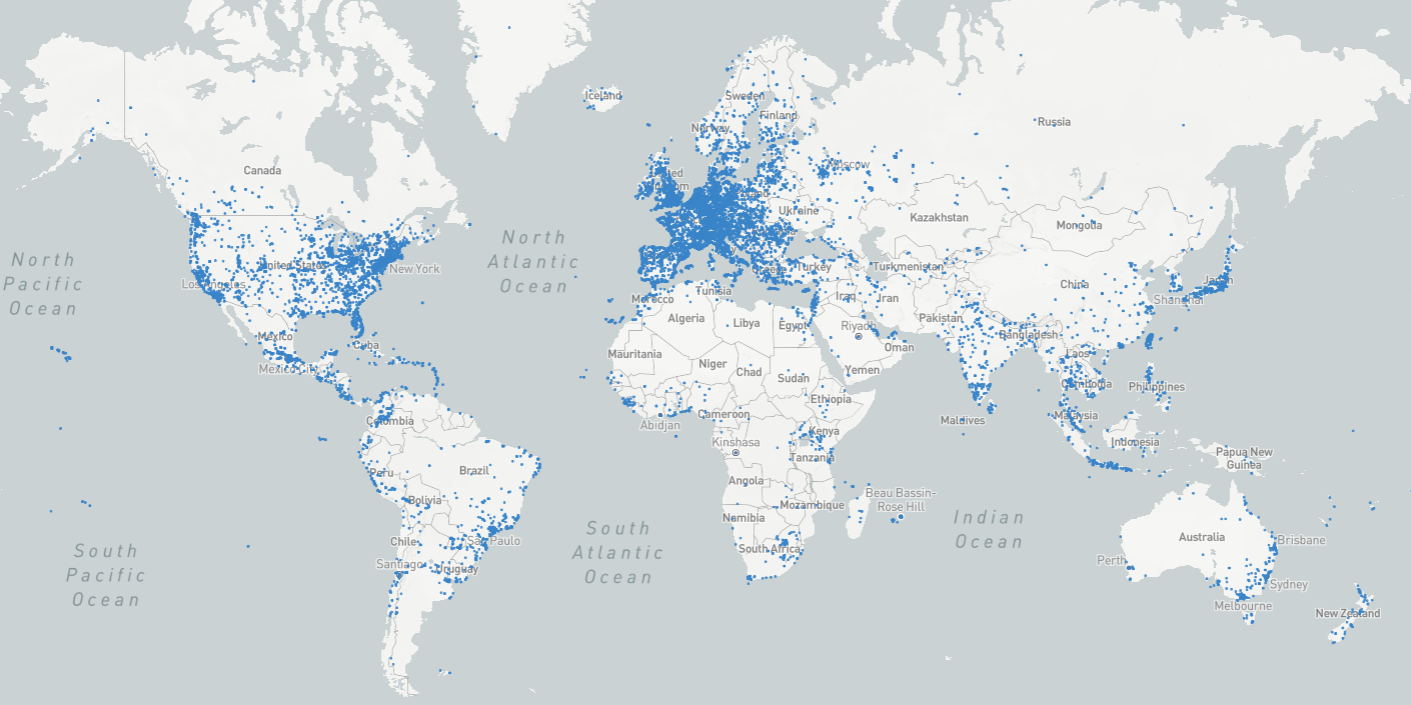"}
    \vspace{-0.15cm}
    \caption{The geographical distribution of images in our dataset.}
    \label{fig:images_distribution}
    \vspace{-0.5cm}
\end{figure}

\vspace{-0.25cm}
\subsection{Pre-filtering}
\label{sec.prefilter}
\vspace{-0.1cm}

In order to maximize the significance of the VIPPGeo dataset for the country recognition task, we applied the following filtering strategies.

\textbf{Remove non--urban images}: we focused exclusively on urban images. Our original research question, in fact, was to analyse if it is possible to recognize a country given the architectural, linguistic, societal and other human-oriented details contained in the picture. Natural scenery images do not contain such details and, in general, are less useful for the country recognition task. Indoor images might also be misleading, given that they may contain design choices that are not necessarily representative of a specific country or region. In order to implement this filter, we used the Places365-CNNs model \cite{zhou2017places}. In particular, Places365 classifies the images according to 365 types of scenery, which can be further categorized as indoor, natural (outdoor natural) and urban (outdoor man-made). To collect the urban images for our dataset, we used the trained ResNet model of Place2 dataset~\cite{zhou2017places} to get the top-5 predicted classes. Then, we summed the prediction probabilities of the urban classes contained in the top-5 list. Finally, we retained only the images with an overall urban probability greater than 0.5.

\textbf{Remove grey-level and outdated images}: as a further filter, we decided to discard grey-level images. The first reason for doing that was to enforce a certain uniformity across the dataset. A second, more compelling, reason is that the features we are interested in, mostly related to human, societal and architectural factors, vary across history. This, in turn, would change significantly the urban aspect of a country, making it difficult to distinguish the peculiar features of a country over time. Given that grey-level images {\em tend} to be old ones, we decided to discard them. For a similar reason, we decided to remove images older than 2012.



\textbf{Remove images portraying {\em mostly} faces}: upon inspection of the images we downloaded, we found that several of them are mostly occupied by one, or more, faces. Clearly, this kind of images are not very relevant for country recognition, hence we decided to discard them. In particular, we removed the images wherein faces occupy more than 10\% of the image area. We did so by using the dlib library \cite{king2009dlib}
to detect and localize faces in the images. 

\textbf{Remove, specific, non-relevant, image classes:} we also found that many of the images we have downloaded focus on specific, non-relevant, subjects like ships in the sea, airfields, airplanes in the sky, and details of sports stadiums. We can argue that these kinds of images are not suitable for country recognition. For instance, an aircraft flying in the sky belonging to a certain airline could be anywhere in the world and therefore is not representative of a country's architectural or semantic style. The same applies to ships. Additionally, sports stadiums such as soccer fields, raceways, and race circuits are hard to classify, unless they contain very specific iconic spots.

In total, after applying the filtering strategies described above, we reduced the number of images from 12 million to 3,813,651.
%

\vspace{-0.25cm}
\subsection{Dataset composition}
\vspace{-0.1cm}
The dataset contains a total of 3,813,651 Images. Each image in the dataset, comes with the corresponding GPS location and country. In particular, the dataset contains 1,716,636 images downloaded from Mapillary, 324,018 from Unsplash, and 1,772,997 from Flickr. The images come from a total of 243 different countries, including very small countries, like Vatican City, or Principality of Monaco.

\begin{table*}[h]
	\centering
	\resizebox{0.85\textwidth}{!}{
	\renewcommand{\arraystretch}{1.25}
			\begin{tabular}{@{}l|c|c|c|c|c|c|c|c|c|c|c|c|c|c|cc@{}}
				\toprule
				{Test sets}&\multicolumn{5}{c|}{\textit{Im2gps - (102 images)}}&\multicolumn{5}{c|}{\textit{Im2gps3k - (1001 images)}} & \multicolumn{5}{c}{\textit{VIPPGeo test set - (76269 images)}}\\
				\midrule
				&Top-1&Top-3&Top-5&Top-10&Bal&Top-1&Top-3&Top-5&Top-10&Bal&Top-1&Top-3&Top-5&Top-10&Bal\\
				\midrule
				
			   ISN  state-of-the-art~\cite{muller2018geolocation} &0.80&0.88&0.88&0.91&0.75&0.66&0.76&0.80&0.85&0.51&0.41&0.52&0.57&0.65&0.31
			   				\\
			   ISN trained on VIPPGeo 
			
			    &0.75&0.82&0.84&0.88&0.72&0.60&0.72&0.77&0.82&0.50&0.63&0.73&0.76&0.81&0.55

			   \\
			   				Proposed method
				&\textbf{0.80}&\textbf{0.91}&\textbf{0.97}&\textbf{0.98}&\textbf{0.80}&\textbf{0.67}&\textbf{0.83}&\textbf{0.88}&\textbf{0.94}&\textbf{0.56}&\textbf{0.66}&\textbf{0.81}&\textbf{0.87}&\textbf{0.93}&\textbf{0.60}\\
				
			   \bottomrule
		\end{tabular}%
		}
		\vspace{-0.1cm}
		\caption{Accuracy of our method and the method in \cite{muller2018geolocation} for country classification. 		}
	\label{tab:results_acc}
	\vspace{-0.5cm}
\end{table*}


While the VIPPGeo dataset can be used as is, that is with each image labelled as belonging to one of 243 countries, some of the counties are extremely small ones. For others, the number of images contained in the dataset is very small, for instance, we were able to gather only 231 images from Kenya and 284 images from Libya. For this reason, in our experiments, we found convenient to group the 243 countries into larger classes consisting of nearby countries. In total, we grouped the images into 61 country classes. Each class includes several countries, which we decided to merge based on two main properties/characteristics: number of images and geographic proximity between countries. With regard to the first criterion, we tried to balance as much as possible the number of images in each of the classes, to avoid possible unbalancing issues during training. With regard to the second criterion, instead, we merged countries according to their geographical proximity. For instance, we grouped Vatican City with Italy, and Egypt with Sudan. It goes without saying, that the way we grouped the images into 61 countries is somewhat arbitrary and other choices could be made by researchers that will use the VIPPGeo dataset for their studies. In the end, the class containing the largest number of images is the one with US images, with 829,345 images. The smallest class is a class containing the United Arab Emirates, with only 6,134 images. The geographical distribution of images in our dataset is shown in Figure~\ref{fig:images_distribution}. The VIPPGeo dataset is publicly available at the following link: \url{https://github.com/alamayreh/VIPPGeo_Dataset}.

\vspace{-0.3cm}
\section{DNN-based classifier}
\label{sec:methods}
\vspace{-0.2cm}

\begin{figure}
    \centering
\includegraphics[width=0.92\columnwidth]{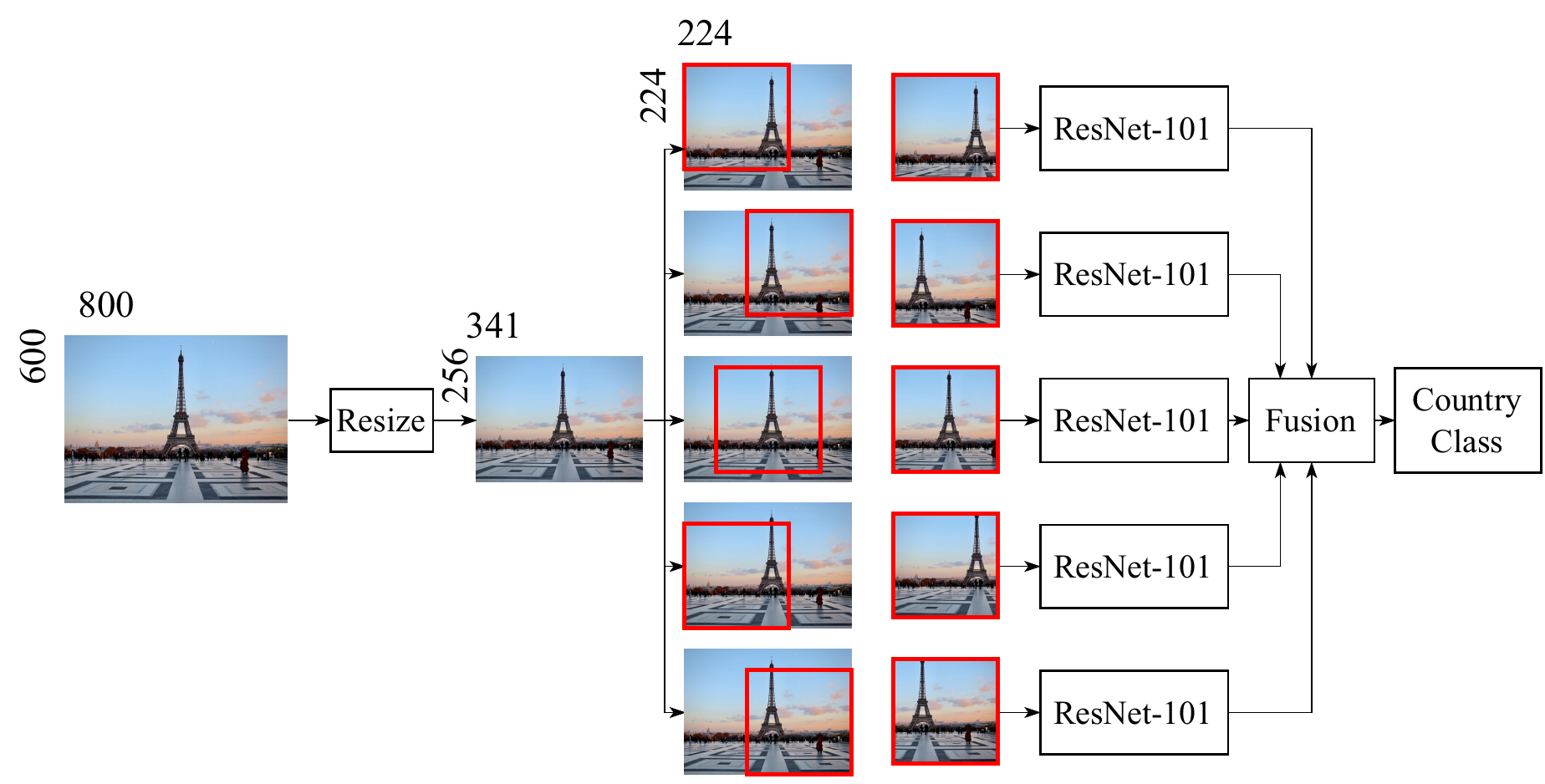}
    \vspace{-0.2cm}
    \caption{Fusion strategy during testing.}
    \label{fig:crops}
    \vspace{-0.6cm}
\end{figure}

The country classifier we have built is based on the ResNet--101 architecture~\cite{he2016deep}. Such architecture has been proven to be extremely successful for the purpose of image classification, and was applied widely in several fields. Given that the size of the network input is fixed ($224\times224$, 3-band images), and given that the images to be classified have very diverse dimensions, we devised a strategy, somewhat similar to that used in \cite{muller2018geolocation}, to analyse the entire image content without changing the aspect ratio of the images, since this could impair the classification. The approach we adopted is sketched in Figure~\ref{fig:crops}. The image is first resized in such a way that the lower dimension is equal to 256. Then, we extract 5 crops of size 224 $\times$ 224: 4 in the corners and 1 in the center. The ResNet--101 classifier is then applied to each crop and the 5 results are fused into a single score. We tried several fusing strategies and found that the best performance is obtained by averaging the scores obtained on the five crops.

%


%

The network analyzing the crops was trained for 25 epochs, starting from pre-trained model on ImageNet. The batch size was set to 320 images. The network was trained by using cross-entropy loss and the Stochastic Gradient Descend (SGD) optimizer with learning rate of $1 \cdot 10^{-2}$ a momentum of 0.9, and a weight decay of $1 \cdot 10^{-4}$.

To avoid overfitting and to facilitate learning robust features, we applied geometric augmentation.
First, we randomly applied horizontal and vertical flipping. After that, we considered a random crop covering at least 2/3 of the image area. Then, we resized the crop to 224 $\times$ 224. Finally, The images were normalized to obtain the mean pixel values and the standard deviation equal to those of Imagenet dataset\footnote{This is common practice when using models pre-trained on Imagenet.}. Training was carried out on the VIPPGeo dataset, which we split into training, validation and testing subsets, according to the following percentages: $96\%$, $2\%$ and $2\%$ (being our dataset large, with this splitting, the test set contains 76,269 images that are enough for testing). Splitting was applied to each country separately. Even if we did our best to balance the number of images in the 61 country classes, balancing is far from perfect. Therefore we had to take proper countermeasures to avoid that the network always decides in favour of the most represented countries. Following~\cite{cui2019class}, we did so by weighting the terms in the loss function according to the inverse of the square root of the class frequency: 

\vspace{-0.25cm}
\begin{equation}
 Loss = - \sum_{y \in \mathcal{D}_{tr}} \sum_{i=1}^{N} \frac{1}{\sqrt{n_i}} y^{i}\log(\bar{y}^{i}),
\end{equation}
where $N$ is the number of classes, $n_i$ is the total number of images in class $i$, $y^{i}$ is the ground truth of image $y$, $\bar{y}^{i}$ is the network score assigned to each class in $N$ for image $y$, and $ \mathcal{D}_{tr}$ is the training set.

\vspace{-0.25cm}
\section{Experiments and Results}
\label{sec:Experiments and Results}
\label{sec:experiments}
\vspace{-0.25cm}

In this section, we report the results of the experiments we performed to assess the effectiveness of the proposed classifier trained on the VIPPGeo dataset. We also report some examples to get some hints about the plausibility of the analysis carried out by the classifier.

The third row of Table~\ref{tab:results_acc} shows the accuracy of our network when tested over the images taken from the Im2GPS~\cite{im2gps}, the Im2GPS-3k dataset~\cite{im2gps_2}, and the test subset of VIPPGeo (76,269 images). The same filtering strategies described in Section \ref{sec.prefilter} were applied to the images in the Im2GPS and Im2GPS-3k datasets. In this way, the 270 images in Im2GPS were lowered down to 102 and the 3,000 images in Im2gps3K were reduced to 1,001. All test images represent urban views in order to have a fair comparison with the VIPPGeo dataset. Top-1, 3, 5, 10 and balanced accuracies \cite{brodersen2010balanced} are reported in the table. Results regarding top-3 accuracy and balanced accuracy are particularly good, taking into account the difficulty of the task and the diversity of the conditions under which the network was tested. Top-1 accuracy shows also good performances, considering the large number of classes involved in the classification. 

We compared the results of the proposed method with those achieved by the method described in \cite{muller2018geolocation}, which is a widely recognized benchmark in the field. Such a method returns the estimation of the GPS location where the image has been shot. For a fair comparison, then, we transformed the geo-coordinates into a country class by mapping the output GPS location returned by ~\cite{muller2018geolocation} to the corresponding country and then to the country class. As shown in the first row in Table~\ref{tab:results_acc}, \cite{muller2018geolocation} performs well on the Img2ps--102 and the Img2ps3k--1001 datasets, but does not generalize well to the images in the VIPPGeo dataset. On the contrary, our method provides comparable or even better performance than \cite{muller2018geolocation} even when tested on images belonging to the datasets used to train  \cite{muller2018geolocation}. 

As a further investigation, we trained the network described in~\cite{muller2018geolocation} on the VIPPGeo dataset obtaining the results reported in the second row of Table~\ref{tab:results_acc}. As we can see, by training the system with the large number of images made available by VIPPGeo, we can boost the performance of~\cite{muller2018geolocation} significantly, thus confirming the importance of the work we made to build the VIPPGeo dataset. Still, upon comparison with the results in Table \ref{tab:results_acc}, the superior performance of our newly proposed method is confirmed.

\begin{table}[ht]
	\centering
	\resizebox{0.99\columnwidth}{!}{
	\renewcommand{\arraystretch}{1.25}
			\begin{tabular}{@{}l|c|c|c|c|c|c|c|c|cc@{}}
				\toprule
				{Test sets}&\multicolumn{3}{c|}{\textit{Imgps - Imgps3k
				}}&\multicolumn{3}{c|}{\textit{Imgps3k}} & \multicolumn{3}{c}{\textit{VIPPGeo test}}\\
				{}&\multicolumn{3}{c|}{\textit{102 images}}&\multicolumn{3}{c|}{\textit{1001 images}} & \multicolumn{3}{c}{\textit{76269 images}}\\
				\midrule
				&Top-1&Top-5&Bal&Top-1&Top-5&Bal&Top-1&Top-5&Bal\\
				\midrule
				
			   Upper left crop &0.73&0.93&0.69&0.59&0.85&0.50&0.57&0.83&0.50
			   \\
			   Upper right crop &0.73&0.94&0.74&0.60&0.84&0.50&0.57&0.82&0.49
			   \\
			   Lower left crop 
			   &0.69&0.90&0.64&0.58&0.84&0.49&0.60&0.83&0.53
			   \\
			   Lower right crop 
			   &0.68&0.89&0.67&0.59&0.84&0.47&0.59&0.83&0.52
			   \\
			   Central crop 
			   &0.72&0.95&0.69&0.60&0.84&0.51&0.60&0.84&0.54
			   \\
			   Max fusion
			   &0.75&0.97&0.78&0.66&0.88&0.55&0.65&0.86&0.59
				\\
				Resize to 224
			    &0.72&0.95&0.65&0.62&0.87&0.51&0.64&0.86&0.60
				\\
			     Averaging  	&\textbf{0.80}&\textbf{0.97}&\textbf{0.80}&\textbf{0.67}&\textbf{0.88}&\textbf{0.56}&\textbf{0.66}&\textbf{0.87}&\textbf{0.60}\\
				
			   \bottomrule
		\end{tabular}%
}
		
		\vspace{-0.25em}
		\caption{Accuracy for different fusion strategies.} 
		
		\vspace{-0.35cm}
	\label{tab:results_crops}
\end{table}

			
				

\vspace{-0.25cm}
\subsection{Ablation study}
\vspace{-0.1cm}

To understand the impact of the way we fuse the results obtained on the five crops on the final accuracy, we carried out some experiments by adopting several fusion strategies.
Specifically, we considered two alternative ways to take into account the entire image content, despite the fixed size of the network input.~\MB{Modify according to new table}~\OMRAN{Done}The first strategy takes the maximum output resulting from the five crops and the second resizes the input image to 224 $\times$ 224 (thus modifying the aspect ratio) and classifies directly the image in one single step. The last three rows of Table~\ref{tab:results_crops} show the obtained results. As we can see, averaging over the five crops gives slightly better performance compared to the other strategies, even if the advantage is a minor one. We also verified that fusing the results of different crops is indeed helpful. We did so by looking at the accuracy obtained by classifying the images based on the result of one single crop. The results obtained, (see rows 1 to 5 of Table~\ref{tab:results_crops}~\BTcomm{why row 1 and not also 2-5?}),~\OMRAN{Done}show clearly that classifying the images based on the partial content contained in one single crop provides significantly worse results.

\vspace{-0.25cm}
\subsection{Plausibility analysis}
\vspace{-0.1cm}

\begin{figure}
    \centering
    \includegraphics[width=0.475\textwidth]{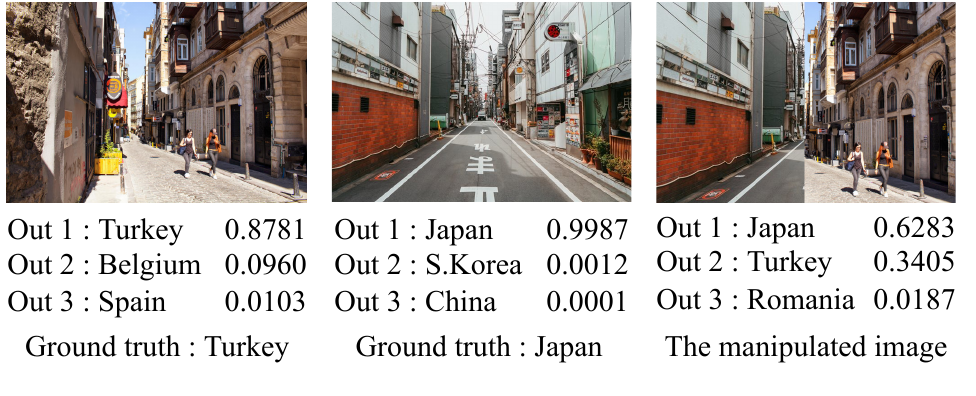}
    \vspace{-0.4cm}
    \caption{Example of Top-3 results provided by the network when two images representing different countries are put together to form a composite image. \BTcomm{Why do we have so much space between the figure and the caption?}~\OMRAN{Done}}
    \label{fig:two_countries}
    \vspace{-0.075cm}
\end{figure}

\begin{figure}
    \centering
\includegraphics[width=0.475\textwidth]{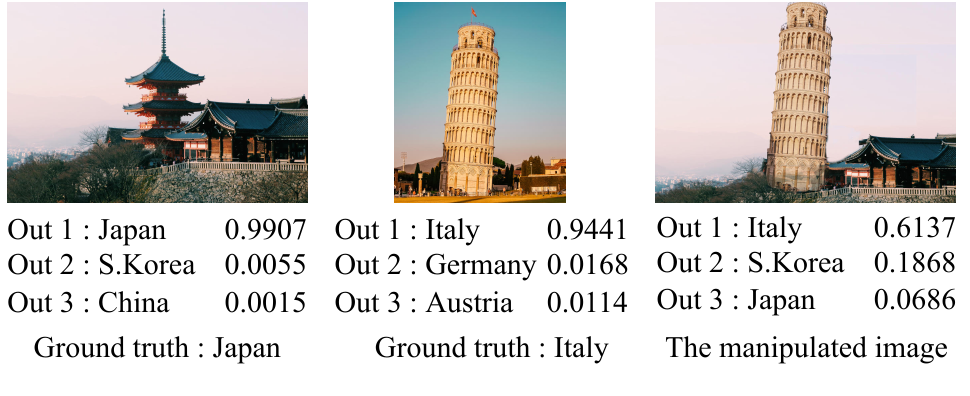}
    \vspace{-0.4cm}
    \caption{Insertion of an iconic monument of one country into a picture representing a different country. The network identifies the two contributing countries in the first positions.}
    
    \label{fig:landmarks_in_countries}
    \vspace{-0.45cm}
\end{figure}

To get some insights about the explainability of the analysis carried out by the network, and make sure that the classification is based on the semantic content of the images, we fed the network with images wherein an iconic monument of one country is pasted into a picture taken from a different country. We also fed the network with images composed of two sub-images taken from different countries. In these cases, we expect the network to be \textit{``confused''}, but still capable of assigning the highest probabilities to the two countries that contributed to form the image. This is indeed the case in many situations, as shown in the examples reported in Figures~\ref{fig:two_countries} and ~\ref{fig:landmarks_in_countries}. In all of the cases, countries which contributed mostly to form the picture are ranked in the first position.
\vspace{-0.3cm}
\section{Conclusions}
\label{sec:conclusions}
\vspace{-0.2cm}

We presented two contributions towards the development of an automatic method for country recognition. Firstly, we introduced a new benchmark dataset, which we named VIPPGeo. The dataset is composed of almost 4 million urban images and was collected to overcome the lack of benchmarks on which deep learning methods can be trained and tested. Secondly, we proposed a novel classification method for automatic country recognition. Differently from state-of-the-art, we casted the problem as a multi-class classification task, instead of the more \textit{``classical''} approach in which the GPS coordinate of the image is first estimated, and then used to trace back to the country where the image was taken. Promising performances were shown in our experiments, both on the new VIPPGeo dataset and other benchmark datasets, outperforming state-of-the-art results. Further perspective may imply the development of novel DL architectures, to recognise not only countries but also cities, towns and villages.

	\vfill\pagebreak
	

	\bibliographystyle{IEEEbib}
	\bibliography{refs}
	
\end{document}